\documentclass[11pt]{article}

\usepackage[T1]{fontenc}
\usepackage{lmodern}
\usepackage{microtype}
\usepackage[margin=1in]{geometry}
\usepackage{setspace}
\usepackage{amsmath}
\usepackage{amssymb}
\usepackage{dsfont}
\usepackage{algorithm}
\usepackage{algpseudocode}
\usepackage{float}

\usepackage{graphicx}
\graphicspath{
    {figures/main/}
    {figures/supplementary/}
}

\usepackage{booktabs}
\usepackage{tabularx}
\usepackage{multirow}
\usepackage{threeparttable}
\usepackage{authblk}

\usepackage{siunitx}
\usepackage[authoryear,round]{natbib}

\usepackage[hidelinks]{hyperref}
\usepackage[nameinlink,noabbrev]{cleveref}

\usepackage{caption}
\usepackage{subcaption}

\usepackage{placeins}

\usepackage{enumitem}

\newcommand{\safeinput}[1]{%
    \IfFileExists{#1.tex}
        {\input{#1}}
        {\typeout{MANUSCRIPT BUILD: #1.tex not yet present}}%
}

\title{}
\author{}
\date{}

\title{Extreme Low-State Quantized Transformer for Electricity-Load Forecasting}

\author{
Muhammad Sabbir Alam\textsuperscript{1},
Harindra Sandun Mavikumbure\textsuperscript{2},
Victor Cobilean\textsuperscript{2},
Milos Manic\textsuperscript{2},
Jayasimha Atulasimha\textsuperscript{1,3,*}
\\
\small \textsuperscript{1}Department of Mechanical and Nuclear Engineering, Virginia Commonwealth University, Richmond, VA, 23284, USA
\\
\small \textsuperscript{2}Department of Computer Science, Virginia Commonwealth University, Richmond, VA, 23284, USA
\\
\small \textsuperscript{3}Department of Electrical and Computer Engineering, Virginia Commonwealth University, Richmond, VA, 23284, USA
\\
\small \textsuperscript{*}Corresponding author: jatulasimha@vcu.edu
}

\date{}

\begin{document}

\maketitle

\begin{abstract}
Aggressive low-state weight quantization can substantially reduce the representational flexibility of neural networks, making it difficult to preserve predictive performance when only a few discrete weight states are available. Here, we investigate this regime for Transformer-based electricity-load forecasting using 3-, 5-, 7-, and 16-state weight quantization across feed-forward-only and attention-plus-feed-forward quantization. Two forecasting tasks are considered, predicting either the next hour or the next 24 hours from the preceding 48 hours, across two co-primary electricity-consumption datasets. Direct post-training quantization (PTQ) increased mean normalized root-mean-square error (NRMSE) relative to full precision by 6.25 and 7.04 percentage points in Datasets A and B, respectively, whereas quantization-aware fine-tuning (QAT-FT) recovered approximately 90.2\% and 94.5\% of these penalties. Quantization-aware training directly from random initialization (Scratch-QAT) also produced competitive low-state solutions without requiring a pretrained full-precision starting point, with particularly favorable behavior in the three-state regime. We further introduce accumulated-update thresholding, which accumulates optimizer-proposed target-weight changes before committing them to the continuous shadow weights. Relative to exactly paired Scratch-QAT controls, this reduced committed shadow-weight updates by approximately 99.7\% and discrete quantized state transitions by approximately 77--80\%, while mean NRMSE changed by only +0.042 and +0.197 percentage points in Datasets A and B. These results demonstrate that useful Transformer forecasting can be retained under highly discrete weight representations while strongly suppressing committed and discrete target-state activity, highlighting a promising route toward edge-based load forecasting in which multilevel synaptic devices could natively encode low-state weights and support compact, low-activity, resource-efficient learning and inference.
\end{abstract}

\section{Introduction}
Deploying neural networks under severe numerical constraints requires more than simply reducing parameter precision after training. When weights are restricted to only a few discrete states, the representational flexibility available to the model becomes strongly limited, making it increasingly difficult to preserve full-precision behavior. This aggressive low-state regime is particularly relevant when model parameters must ultimately be represented by a small number of accessible levels, but it also poses a fundamental learning question: how much predictive capability can be retained when the weight space itself is highly discrete? Addressing this question requires training strategies designed for low-state operation rather than treating quantization only as a final compression step \citep{Hubara2018,esser2020lsq,Nagel2021,lin2024awq,shao2024omniquant}. Recent work continues to show that preserving accuracy becomes particularly demanding in highly constrained low-bit Transformer representations \citep{lin2024awq,shao2024omniquant}.

Post-training quantization (PTQ) is attractive because it can reduce weight precision without modifying the original training procedure or requiring end-to-end retraining. At moderate precision, this simplicity can make PTQ an efficient route from a pretrained full-precision model to a lower-precision representation, and improved rounding, reconstruction, and calibration methods have extended its reach toward lower bit widths \citep{nagel2020adaround,li2021brecq,frantar2023gptq,xiao2023smoothquant,lin2024awq,shao2024omniquant}. However, the difficulty becomes much greater when the available weight space collapses to only a few discrete states, where coarse projection can substantially disturb the learned solution. Quantization-aware training (QAT) addresses this limitation by incorporating the low-state representation into training itself, allowing the model to adapt while quantization is active \citep{Krishnamoorthi2018,Jacob2018,esser2020lsq,Nagel2021}.

When such quantization is incorporated during training, a common pathway is quantization-aware fine-tuning (QAT-FT), which begins from a pretrained full-precision solution and adapts it to the discrete weight space \citep{Nagel2021}. This raises a different question from whether QAT can recover PTQ-induced degradation: must a useful extremely low-state model originate from a full-precision precursor at all? Prior studies show that neural networks can be optimized directly with very low-bit weights and activations \citep{Hubara2018,esser2020lsq}. If competitive solutions can emerge directly from random initialization while quantization is active from the outset, low-state learning need not be viewed only as adaptation of an already trained continuous model. Direct random-initialized QAT (Scratch-QAT) therefore provides a complementary test of whether the constrained weight space itself can support effective learning without any pretraining.

Low-state learning also raises an issue not captured by forecasting accuracy alone: how often training actually requires a committed parameter change or a transition between discrete weight states. This distinction is relevant to in-memory and synaptic-device computing, where discrete weights can be represented by programmable conductance states and physical reprogramming carries implementation costs \citep{legallo2018mixedprecision,sebastian2020memory,azam2020voltage,Misba2022,alam2024quantizedautoencoder,dhull2025quantizeddw,sehgal2025triplelevel,Aguirre2024,jiang2026memristor}. Mixed-precision computational-memory work has further shown that small numerical updates can be accumulated at higher precision and transferred to device states only when sufficiently large relative to programming granularity \citep{nandakumar2020mixedprecision}. We therefore separate optimizer-proposed updates from committed weight updates and actual discrete state transitions, and ask whether small adjustments can be accumulated before commitment. Such a strategy could reduce state-changing activity while preserving gradual optimization, with prospective relevance to multilevel synaptic devices and resource-constrained edge hardware \citep{sebastian2020memory,alam2024quantizedautoencoder,dhull2025quantizeddw,bettayeb2024memristor}.

These questions are particularly relevant to electricity-load forecasting, where models must capture recurring temporal structure and short-term variability over near-term and multi-step horizons \citep{HongFan2016,amarasinghe2017deep,almisba2023spintronicload}. Transformer and attention-based architectures have become prominent in time-series forecasting, with recent developments targeting interpretable multi-horizon prediction, long-sequence efficiency, decomposition, patch-based representations, and variable-centric modeling \citep{lim2021tft,zhou2021informer,wu2021autoformer,zhou2022fedformer,nie2023patchtst,liu2024itransformer,su2025transformerreview}. Transformer-based methods have also been applied directly to electricity-load forecasting \citep{chan2024transformerload}. We therefore investigate aggressive low-state quantization in this setting. The Transformer \citep{Vaswani2017} is well suited to this study because its self-attention mechanism can model temporal dependencies across the input sequence, while its distinct attention and feed-forward weight blocks provide two major parameter groups whose sensitivity to severe quantization can be examined separately. By evaluating both one-hour and 24-hour forecasting across two co-primary electricity-consumption datasets with contrasting load characteristics, this study examines whether extremely low-state Transformer learning remains effective across different forecasting horizons, demand patterns, and quantization scopes.

This study makes five principal contributions. First, it establishes the behavior of Transformer forecasting under aggressive 3-, 5-, 7-, and 16-state weight quantization across feed-forward-only and attention-plus-feed-forward quantization, showing that direct PTQ becomes particularly damaging at the lowest state counts, whereas performance progressively approaches full-precision behavior as state resolution increases. Second, it shows that quantization-aware fine-tuning recovers most of the severe PTQ penalty, recovering approximately 90.2\% in Dataset A and 94.5\% in Dataset B. Third, it demonstrates that training directly from random initialization under quantized forward conditions---Scratch-QAT---can also produce competitive low-state solutions without requiring a pretrained full-precision starting point, with particularly favorable behavior in the most restrictive three-state regime. Fourth, it introduces accumulated-update thresholding, in which small optimizer-proposed weight changes are accumulated before being committed to the target weights; only committed changes can subsequently alter the corresponding quantized weight from one discrete state to another. This approach reduces committed weight updates by approximately 99.7\% and discrete state transitions by approximately 77--80\% while producing only small mean changes in forecasting error, demonstrating that optimizer activity can be strongly decoupled from the weight changes that are actually committed and from those that change the discrete state representation. Finally, the principal PTQ, QAT-FT, and thresholding effects reproduce across the two co-primary datasets and all four distinct load-series analysis units, including transfer of the methodology to Dataset B without Dataset-B-specific retuning, supporting the robustness of the observed low-state learning behavior across heterogeneous electricity-demand settings.

The remainder of this paper is organized as follows. Section 2 describes the datasets, forecasting protocol, quantized Transformer framework, training strategies, and evaluation methods. Section 3 presents the main forecasting, quantization, thresholding, and cross-dataset results. Section 4 discusses the implications, limitations, and prospective hardware relevance of the findings. Section 5 concludes the paper.

\section{Methods}

\subsection{Co-primary datasets and forecasting tasks}

Two publicly available electricity-consumption datasets from different geographical and consumption settings were treated as co-primary datasets and evaluated using the same forecasting formulation (Table~\ref{tab:datasets_protocol}).

\begin{table}[t]
\centering
\caption{Co-primary datasets and forecasting protocol.}
\label{tab:datasets_protocol}

\small
\begin{tabularx}{\textwidth}{@{}p{0.22\textwidth}XX@{}}
\toprule
 & Dataset A & Dataset B \\
\midrule

Source &
UCI ElectricityLoadDiagrams20112014 &
UCI Individual Household Electric Power Consumption \\

Role &
Co-primary &
Co-primary dataset \\

Series &
MT\_004, MT\_022, MT\_324 &
Household\_Power \\

Distinct series &
3 &
1 \\

Modeling resolution &
Hourly mean from 15-minute measurements &
Hourly \\

Study period &
2012-01-01 through 2014-12-31 &
2007-01-01 through 2009-12-31 \\

Input window &
48 h &
48 h \\

Forecast tasks &
H1 (48$\rightarrow$1 h), H24 (48$\rightarrow$24 h) &
H1 (48$\rightarrow$1 h), H24 (48$\rightarrow$24 h) \\

H24 strategy &
Direct multi-output forecasting &
Direct multi-output forecasting \\

Final training period &
2012-01-01 through 2014-06-30 &
2007-01-01 through 2009-06-30 23:00 \\

Final test period &
2014-07-01 through 2014-12-31 &
2009-07-01 through 2009-12-31 \\

Final sample counts &
H1 train = 21,693; H1 test = 4,416; H24 train = 21,601; H24 test = 4,393 &
H1 train = 19,558; H1 test = 4,059; H24 train = 18,556; H24 test = 3,875 \\

Data-validity handling &
Main modeling series not interpolated; samples touching identified DST artifact hours were excluded. &
An hourly value is valid only when exactly 60 valid minute-level Global\_active\_power measurements are present. Invalid hours are retained as NaN. No interpolation or imputation is performed. Forecast samples are retained only when all 48 input hours and all target hours are finite. \\

\bottomrule
\end{tabularx}
\end{table}

Dataset A was the UCI \textit{ElectricityLoadDiagrams20112014} dataset \citep{Trindade2015}, containing measurements from 370 clients in Portugal. Three anonymous client series, MT\_004, MT\_022, and MT\_324, were analyzed from 1 January 2012 through 31 December 2014. The original 15-min measurements were averaged to hourly resolution. The three series were selected to provide contrasting temporal and statistical load characteristics; detailed descriptive characterization is reported in the Supplementary Information. No interpolation was applied, and forecasting samples intersecting the three identified daylight-saving-time artifact hours were excluded.

Dataset B was the UCI \textit{Individual Household Electric Power Consumption} dataset \citep{HebrailBerard2006}. Global\_active\_power from a single household in Sceaux, France, was analyzed from 1 January 2007 through 31 December 2009. Minute-level measurements were averaged to hourly resolution, with an hour considered valid only when all 60 minute-level observations were available. Invalid hours were retained as missing values, no interpolation or imputation was performed, and forecasting samples were retained only when all required input and target observations were finite.

Both datasets used the preceding 48 hourly observations as input. For the one-hour-ahead task (H1),

\begin{equation}
[x_{t-47},\ldots,x_t]\rightarrow x_{t+1}.
\label{eq:h1_forecast}
\end{equation}

whereas for the day-ahead task (H24),
\begin{equation}
[x_{t-47},\ldots,x_t]
\rightarrow
[x_{t+1},\ldots,x_{t+24}].
\end{equation}

H24 was therefore implemented as a direct multi-output forecast, rather than by recursively feeding earlier predictions back into the model. Chronological development, normalization, and final-test procedures are described in Section~2.2.
\subsection{Development protocol and final-test locking}

A strictly chronological development protocol separated model development from final performance evaluation. Each co-primary dataset used three expanding-window validation folds, with every validation interval occurring later than its corresponding training interval (Table~\ref{tab:development_protocol}). Final test periods were reserved until completion of development and methodological locking.

\begin{table}[t]
\centering
\caption{Chronological development and final evaluation protocol}
\label{tab:development_protocol}
\small
\begin{tabularx}{\textwidth}{@{}llXX@{}}
\toprule
Dataset & Stage & Training period & Validation / test period \\
\midrule

Dataset A
& Fold 1
& Jan 2012--Dec 2012
& Jan 2013--Jun 2013 \\

& Fold 2
& Jan 2012--Jun 2013
& Jul 2013--Dec 2013 \\

& Fold 3
& Jan 2012--Dec 2013
& Jan 2014--Jun 2014 \\

& Final
& Jan 2012--30 Jun 2014
& 1 Jul--31 Dec 2014, final test \\

\addlinespace

Dataset B
& Fold 1
& Jan 2007--Dec 2007
& Jan 2008--Jun 2008 \\

& Fold 2
& Jan 2007--Jun 2008
& Jul 2008--Dec 2008 \\

& Fold 3
& Jan 2007--Dec 2008
& Jan 2009--Jun 2009 \\

& Final
& Jan 2007--30 Jun 2009
& 1 Jul--31 Dec 2009, final test \\

\bottomrule
\end{tabularx}
\end{table}

Normalization was performed independently within each development fold using z-score scaling fitted only to the corresponding training interval. Validation observations were excluded from scaler estimation. The same training-only normalization principle was used for final model fitting; invalid or excluded observations were handled according to the dataset-specific validity rules described in Section~2.1.

After Dataset-A development, the principal experimental choices were locked before final-test evaluation. These included the forecasting tasks, horizon-specific Transformer configurations, 3-, 5-, 7-, and 16-state representations, two quantization locations, QAT-FT and Scratch-QAT procedures, Scratch training endpoints, and the selected high-$\alpha$ schedule. All selections were based on development-stage evidence rather than final-test performance.

The methodology established using Dataset A was subsequently transferred to Dataset B without Dataset-B-specific retuning. State counts, quantization locations, the high-$\alpha$ rule, Transformer configurations, and QAT/Scratch-QAT procedures were not reselected using Dataset-B performance. Following final-test access in either dataset, no performance-driven retraining or modification of these methodological choices was performed.
\subsection{Controlled Transformer architecture}

A conventional encoder-only Transformer was used as the common forecasting backbone for both co-primary datasets. The architecture was intentionally kept relatively compact so that the experiments could isolate the effects of low-state weight quantization and training strategy, rather than introduce a specialized forecasting architecture as an additional source of variation. The same frozen horizon-specific Transformer configurations were used for Dataset A and Dataset B, consistent with the methodology-transfer protocol described in Section 2.2.

Each input sample had shape \(48\times1\), corresponding to the preceding 48 hourly load observations. The scalar value at each time step was first projected through a dense layer to a 64-dimensional representation (\(d_{\mathrm{model}}=64\)), after which a trainable positional embedding of length 48 and dimension 64 was added to preserve temporal order. The resulting sequence was processed by pre-normalization Transformer encoder blocks. Each block contained four-head self-attention, with an attention key dimension of \(64/4=16\) per head, followed by a position-wise feed-forward network of dimension \(64\rightarrow128\rightarrow64\) with GELU activation. Layer normalization preceded both the self-attention and feed-forward sublayers, and residual connections were applied after each sublayer. Dropout was applied within the attention and feed-forward paths.

After the final encoder block, a final LayerNorm was applied and only the representation corresponding to the most recent input time step was retained for forecasting. This 64-dimensional last-token representation was passed to a dense output layer containing either one unit for H1 or 24 units for H24, producing the direct forecasts defined in Section 2.1. The H1 configuration used three encoder blocks, dropout of 0.05, and 103,809 trainable parameters, whereas the H24 configuration used one encoder block, dropout of 0.10, and 38,360 trainable parameters. Both configurations retained \(d_{\mathrm{model}}=64\), four attention heads, and an FFN dimension of 128.
\subsection{Low-state weight quantization}

Weight quantization used a uniform symmetric per-weight-tensor quantizer with $L$ discrete states. For each targeted weight tensor $W$,

\begin{equation}
s=\max |W|,
\end{equation}

where $s$ denotes the largest absolute weight value within that tensor. The spacing between adjacent states was

\begin{equation}
\Delta=\frac{2s}{L-1},
\label{eq:quant_spacing}\end{equation}

giving

\begin{equation}
\mathcal{Q}_L=
\{-s,\,-s+\Delta,\ldots,-s+(L-1)\Delta\}.
\end{equation}

Each targeted weight was mapped deterministically to its nearest available state. Because quantization was performed per weight tensor, each targeted kernel had its own symmetric range $[-s,s]$. The reference tensor used to establish this fixed range depended on the training pathway and is described in Section~2.5.

Development-stage experiments evaluated 2, 3, 4, 5, 7, 8, 16, 32, 64, 128, and 256 states. Forecasting performance progressively approached the full-precision reference as state resolution increased, while differences among the higher-state conditions became comparatively small. The final 3-, 5-, 7-, and 16-state set was therefore selected to focus the main experiments on the strongly quantized regime while spanning progressively finer state resolution, with 16 states providing a higher-resolution condition near the region where further increases produced diminishing differences. The odd 3-, 5-, and 7-state symmetric grids additionally contain an exact zero state. Such low-state representations are also relevant to prospective multilevel synaptic-device implementations, where model weights could be encoded using a limited number of accessible physical states.

Two quantization locations were evaluated. In FFN-Q, only the two feed-forward weight kernels within each encoder block were quantized. In Attention+FFN-Q, quantization additionally included the query, key, value, and attention-output kernels. Activations, biases, LayerNorm operations, softmax, and all non-target parameters remained in FP32.

For H1, FFN-Q quantized 49,152 of 103,809 trainable parameters (47.35\%), whereas Attention+FFN-Q quantized 98,304 (94.70\%). For H24, the corresponding values were 16,384 of 38,360 parameters (42.71\%) and 32,768 (85.42\%), respectively. Thus, Attention+FFN-Q imposed the low-state constraint on the large majority of trainable parameters, making it substantially more restrictive than FFN-Q. The same state counts and quantization locations were used for both co-primary datasets.
\subsection{Training strategies}

Four training and quantization pathways were compared: full precision (FP), post-training quantization (PTQ), quantization-aware fine-tuning (QAT-FT), and random-initialized quantization-aware training (Scratch-QAT). All neural-network optimization used Adam with mean-squared error (MSE) loss. PTQ and QAT-FT were independent branches originating from the same pretrained FP checkpoint, whereas Scratch-QAT was initialized randomly and trained with quantization active from the beginning.

For the quantization-aware training pathways, $\alpha$ denotes the dimensionless accumulated-update threshold coefficient controlling when an optimizer-proposed change to a targeted continuous weight is committed. For a targeted tensor with fixed symmetric range $[-s,s]$, the threshold magnitude is $\alpha s$. Thus, $\alpha=0$ represents the immediate-commit condition, whereas $\alpha>0$ allows smaller proposed changes to accumulate before commitment. The complete accumulated-update procedure and selected high-$\alpha$ rule are described in Section~2.6.

\subsubsection{Full precision}

The FP model provided the unquantized reference for each dataset and forecasting horizon. All weights were trained in full precision using the horizon-specific architecture described in Section~2.3. The resulting checkpoint served as both the performance reference and the common starting point for the PTQ and QAT-FT branches.

\subsubsection{Post-training quantization}

For PTQ, the trained FP checkpoint was quantized directly using the state count and quantization location defined in Section~2.4. For each targeted tensor, the fixed symmetric range $[-s,s]$ was obtained from the corresponding trained FP weights, which were then mapped to their nearest available quantization states. No retraining or weight adaptation was performed after quantization.

\subsubsection{Quantization-aware fine-tuning}

QAT-FT began independently from the same pretrained FP checkpoint used for PTQ; it was not initialized from the PTQ model. Per-tensor quantization ranges were calculated from the original FP checkpoint and held fixed during fine-tuning.

Following the straight-through estimator (STE) principle used in low-precision training \citep{Hubara2018}, continuous shadow weights were maintained for each targeted kernel while their quantized counterparts were used for model computation. This distinction is particularly important at very low state counts, where individual optimizer updates can be substantially smaller than the spacing between neighboring quantization states. The continuous shadow representation preserves such fine-grained updates even when the corresponding quantized weight does not immediately change state.

During each minibatch, the targeted kernels were quantized for the forward pass and gradient calculation, after which the continuous shadow weights were restored before the Adam update. Thus, the network was evaluated under the prescribed discrete-state representation while optimization acted on continuous underlying parameters. QAT-FT was performed for 25 epochs with a learning rate of $1\times10^{-4}$, batch size 128, seed 42, and $\alpha=0$. No early stopping or learning-rate scheduler was used.

\subsubsection{Random-initialized quantization-aware training (Scratch-QAT)}

Scratch-QAT tested whether useful low-state solutions could be learned without a pretrained FP starting point. The Transformer was freshly initialized using Glorot-uniform initialization with seed 42, and the resulting initialization was saved as $W_0$. For each targeted tensor, the fixed quantization range $[-s,s]$ was derived from this same $W_0$, rather than from a pretrained FP checkpoint. Quantization was active from the first minibatch, using the same continuous-shadow-weight and STE-style framework as QAT-FT.

Scratch-QAT used Adam with a learning rate of $2\times10^{-3}$, batch size 128, and seed 42. The primary endpoint was 40 epochs (E40), while a 25-epoch checkpoint (E25) was retained for a training-budget-matched comparison with QAT-FT.

Two exactly paired Scratch-QAT conditions were evaluated: Scratch $\alpha=0$ and Scratch high-$\alpha$. Each pair began from the identical saved $W_0$ and used the same optimizer initialization, minibatch ordering, learning rate, batch size, quantization range, state count, quantization location, and epoch budget; only the accumulated-update threshold differed. The thresholding procedure and high-$\alpha$ rule are described in Section~2.6.
\subsection{Accumulated-update thresholding}

Accumulated-update thresholding was introduced within Scratch-QAT to control how frequently optimizer-proposed changes were committed to the targeted continuous shadow weights. The rule operates after Adam has generated a proposed continuous target-weight update but before that change is committed. It therefore acts at the weight-commit stage without removing or skipping conventional forward passes, backward passes, minibatches, or Adam optimizer steps.

For each targeted weight element $i$, a non-trainable accumulator $A_i$ was initialized to zero. Let $w_i^{(t)}$ denote the currently committed continuous shadow weight at minibatch $t$, and let $\widetilde{w}_i^{(t+1)}$ denote the optimizer-proposed continuous value after clipping to the fixed tensor range $[-s,s]$. The proposed target-weight change was

\begin{equation}
\delta w_i^{(t)}
=
\widetilde{w}_i^{(t+1)}
-
w_i^{(t)},
\end{equation}

and was added to the existing accumulator,

\begin{equation}
\bar{A}_i^{(t)}
=
A_i^{(t)}
+
\delta w_i^{(t)}.
\end{equation}

Thus, sub-threshold optimizer proposals were retained rather than discarded and could combine with subsequent proposed changes. The tensor-specific commit threshold was

\begin{equation}
\tau=\alpha s,
\end{equation}

and a target element became eligible for commitment when

\begin{equation}
\left|\bar{A}_i^{(t)}\right|\geq\alpha s.
\end{equation}

The threshold is therefore defined relative to the tensor range $s$, not the quantization-state spacing $\Delta$. For $\alpha=0$, every nonzero proposed target-weight change was committed immediately; for $\alpha>0$, sub-threshold changes remained stored until their accumulated magnitude reached the threshold.

When the criterion was satisfied, the accumulated change was applied to the continuous shadow weight and the corresponding accumulator was reset. Otherwise, the committed shadow weight remained unchanged and the accumulator was retained. The next quantized value was always obtained from the currently committed shadow weight rather than from the uncommitted accumulator. Consequently, a committed continuous shadow-weight update and a discrete quantized state transition are distinct events: a committed update does not necessarily cross a quantization-state boundary.

The selected high-$\alpha$ schedule was related to the geometry of the uniform quantization grid. Since

\[
\Delta=\frac{2s}{L-1}, 
\]

from Eq.~\eqref{eq:quant_spacing}, choosing $\alpha=1/(L-1)$ gives $\alpha s=\Delta/2$. The final high-$\alpha$ rule was therefore

\begin{equation}
\alpha_{\mathrm{high}}(L)
=
\min\left(0.25,\frac{1}{L-1}\right),
\end{equation}

yielding $\alpha=0.25$, $0.25$, $1/6$, and $1/15$ for 3, 5, 7, and 16 states, respectively. The cap at 0.25 prevented the threshold from exceeding the largest initially considered value for the coarsest representations. A moderate-$\alpha$ schedule was also evaluated during Dataset-A development, but the high-$\alpha$ schedule provided stronger activity suppression while retaining comparable forecasting performance and was therefore selected as the primary thresholded condition before final-test evaluation. The same rule was transferred unchanged to Dataset B.

The nonzero threshold exploration was motivated by prior work on low-resolution synaptic and memory implementations and tolerance to weight perturbations \citep{Misba2022,Joshi2020}. Here, however, $\alpha$ specifically denotes the accumulated-update commit-threshold coefficient rather than a directly measured physical-device quantity.

The complete operational sequence is summarized in Algorithm~\ref{alg:accumulated_update}.

\begin{algorithm}[H]
\caption{Accumulated-update thresholding for Scratch-QAT}
\label{alg:accumulated_update}
\begin{algorithmic}[1]

\Require Training minibatches; targeted continuous shadow weights $w$; fixed tensor ranges $[-s,s]$; state count $L$; threshold coefficient $\alpha$

\Statex \textit{Notation:} $k_i^{(t)}$ denotes the discrete quantization-state index associated with targeted weight element $i$ at minibatch $t$

\State Initialize element-wise accumulators $A_i \gets 0$

\State Initialize discrete states
\[
k_i^{(0)}
\gets
\operatorname{index}
\left[
Q_L\left(w_i^{(0)}\right)
\right]
\]

\For{each minibatch $t$}

    \State Save the current continuous target shadow weights

    \State Quantize targeted weights using $Q_L(\cdot)$

    \State Perform the forward pass using the quantized model and compute the MSE loss

    \State Perform backpropagation under the quantized model

    \State Restore the continuous shadow weights

    \State Apply Adam to obtain proposed continuous target weights

    \For{each targeted weight element $i$}

        \State Clip the optimizer-proposed value to the fixed range $[-s,s]$

        \State
        $\displaystyle
        \delta w_i^{(t)}
        \gets
        \widetilde{w}_i^{(t+1)}
        -
        w_i^{(t)}
        $

        \State
        $\displaystyle
        \bar{A}_i^{(t)}
        \gets
        A_i^{(t)}
        +
        \delta w_i^{(t)}
        $

        \If{$
        (\alpha=0 \land \bar{A}_i^{(t)}\neq0)
        \lor
        (\alpha>0 \land |\bar{A}_i^{(t)}|\geq\alpha s)
        $}

            \State
            $\displaystyle
            w_i^{(t+1)}
            \gets
            \operatorname{clip}
            \left(
            w_i^{(t)}
            +
            \bar{A}_i^{(t)},
            -s,s
            \right)
            $

            \State
            $\displaystyle
            A_i^{(t+1)}
            \gets 0
            $

        \Else

            \State
            $\displaystyle
            w_i^{(t+1)}
            \gets
            w_i^{(t)}
            $

            \State
            $\displaystyle
            A_i^{(t+1)}
            \gets
            \bar{A}_i^{(t)}
            $

        \EndIf

        \State Recalculate the discrete quantized state:
        \[
        k_i^{(t+1)}
        \gets
        \operatorname{index}
        \left[
        Q_L
        \left(
        w_i^{(t+1)}
        \right)
        \right]
        \]

        \State Record proposed updates, committed updates, state transitions, and transition distance

    \EndFor

\EndFor

\end{algorithmic}
\end{algorithm}

Proposed updates, committed shadow-weight updates, discrete state transitions, and transition distance were recorded separately; their formal definitions are given in Section~2.7.

\FloatBarrier
\subsection{Update and transition accounting}

To quantify the effect of accumulated-update thresholding independently of forecasting accuracy, training activity was recorded at four levels: proposed target-weight updates, committed shadow-weight updates, discrete quantized state transitions, and transition distance. All quantities were counted for individual elements of the targeted weight tensors and accumulated across minibatches, targeted tensors, and epochs. They therefore represent weight-element activity rather than the number of optimizer steps. Here, $t$ indexes minibatch update steps across all epochs and $i$ indexes targeted weight elements.

A proposed target-weight update was counted whenever Adam generated a nonzero change in a targeted continuous shadow weight after clipping the optimizer-proposed value to its fixed tensor range. Using the proposed change defined in Section~2.6,

\begin{equation}
p_i^{(t)}
=
\mathds{1}
\left[
\delta w_i^{(t)}\neq0
\right],
\end{equation}

and

\begin{equation}
N_{\mathrm{prop}}
=
\sum_t\sum_i p_i^{(t)}.
\end{equation}

A committed shadow-weight update was counted only when thresholding and range clipping produced an actual change in the committed continuous shadow weight:

\begin{equation}
u_i^{(t)}
=
\mathds{1}
\left[
w_i^{(t+1)}
\neq
w_i^{(t)}
\right],
\end{equation}

with

\begin{equation}
N_{\mathrm{commit}}
=
\sum_t\sum_i u_i^{(t)}.
\end{equation}

Thus, crossing the threshold alone was insufficient to count a commit if clipping left the shadow value unchanged. For the $\alpha=0$ control, every nonzero proposed update was committed, and the implementation verified $N_{\mathrm{prop}}=N_{\mathrm{commit}}$.

A discrete quantized state transition was counted separately using the state indices defined in Section~2.6:

\begin{equation}
q_i^{(t)}
=
\mathds{1}
\left[
k_i^{(t+1)}
\neq
k_i^{(t)}
\right],
\end{equation}

giving

\begin{equation}
N_{\mathrm{trans}}
=
\sum_t\sum_i q_i^{(t)}.
\end{equation}

Because a committed continuous shadow-weight change can remain within the same quantization interval,

\begin{equation}
N_{\mathrm{trans}}
\leq
N_{\mathrm{commit}}.
\end{equation}

Committed updates and discrete state transitions therefore represent distinct levels of target-weight activity.

Transition distance additionally captured the magnitude of movement through the discrete state space:

\begin{equation}
D_{\mathrm{trans}}
=
\sum_t\sum_i
\left|
k_i^{(t+1)}
-
k_i^{(t)}
\right|.
\end{equation}

An adjacent-state transition contributes one unit, whereas a transition spanning multiple state intervals contributes correspondingly more. Consequently,

\begin{equation}
D_{\mathrm{trans}}
\geq
N_{\mathrm{trans}}.
\end{equation}

The commit acceptance rate was defined as

\begin{equation}
R_{\mathrm{commit}}
=
100
\frac{N_{\mathrm{commit}}}
{N_{\mathrm{prop}}}
\quad(\%),
\end{equation}

representing the percentage of nonzero optimizer-proposed target-weight changes that resulted in actual committed shadow-weight changes.

Because the number of targeted parameters differs across quantization locations and forecasting architectures, activity reductions were evaluated within exactly paired high-$\alpha$ and $\alpha=0$ Scratch-QAT runs rather than by comparing unrelated raw counts. These paired runs used identical target tensors, initialization, minibatch sequence, and training budget, as described in Sections~2.5--2.6.

These activity measures do not represent reductions in conventional optimization workload. The paired $\alpha=0$ and high-$\alpha$ runs executed the same minibatches, forward/backward evaluations, and Adam optimizer steps. Thresholding instead changes how frequently proposed target-weight changes are committed and how frequently those commits produce discrete state transitions. Such reductions may be relevant to prospective implementations in which physical state changes incur write costs; device write energy, endurance, lifetime, latency, and physical programming pulses were not measured here.
\subsection{Evaluation metrics, baselines, and evidence hierarchy}

Forecasting performance was evaluated after inverse transformation to the original load scale using root-mean-square error (RMSE), mean absolute error (MAE), and normalized root-mean-square error (NRMSE). For $N$ observations $y_n$ and predictions $\hat y_n$,

\begin{equation}
\mathrm{RMSE}
=
\sqrt{
\frac{1}{N}
\sum_{n=1}^{N}
(\hat y_n-y_n)^2
},
\end{equation}

\begin{equation}
\mathrm{MAE}
=
\frac{1}{N}
\sum_{n=1}^{N}
|\hat y_n-y_n|,
\end{equation}

and

\begin{equation}
\mathrm{NRMSE}
=
100
\frac{\mathrm{RMSE}}
{\bar y}
\quad(\%),
\qquad
\bar y
=
\frac{1}{N}
\sum_{n=1}^{N}y_n.
\end{equation}

NRMSE was the principal forecasting metric because it expresses error relative to the mean load level and facilitates comparison among series with different absolute magnitudes. RMSE and MAE were retained as complementary measures.

For H1, metrics were calculated over all valid next-hour predictions. For H24, the primary metrics were calculated jointly over all valid samples and all 24 simultaneously predicted forecast hours. Metrics were also calculated separately for forecast hours 1--24 to examine lead-time-dependent behavior, with those trajectories reported in the results section.

Three conventional baselines were used for context: persistence, daily seasonality, and weekly seasonality. Persistence repeated the most recently observed load value across the required forecast horizon. The daily and weekly seasonal baselines used

\begin{equation}
\hat y_t=y_{t-24}
\end{equation}

and

\begin{equation}
\hat y_t=y_{t-168},
\end{equation}

respectively. For H24, these references were constructed separately for each target timestamp. The weekly baseline was treated as a strong seasonal benchmark rather than an information-matched comparator because it accesses observations outside the Transformer's fixed 48-h input window.

Differences in NRMSE are reported in percentage points (pp). For example, the PTQ penalty relative to FP was

\begin{equation}
\Delta_{\mathrm{PTQ-FP}}
=
\mathrm{NRMSE}_{\mathrm{PTQ}}
-
\mathrm{NRMSE}_{\mathrm{FP}}.
\end{equation}

The fraction of the PTQ-associated penalty recovered by QAT-FT was defined as

\begin{equation}
R_{\mathrm{QAT}}
=
100
\frac{
\mathrm{NRMSE}_{\mathrm{PTQ}}
-
\mathrm{NRMSE}_{\mathrm{QAT\text{-}FT}}
}{
\mathrm{NRMSE}_{\mathrm{PTQ}}
-
\mathrm{NRMSE}_{\mathrm{FP}}
}.
\end{equation}

Thus, $R_{\mathrm{QAT}}=100\%$ denotes recovery to the FP NRMSE level, whereas values below 100\% indicate a residual gap relative to FP.

The broad Dataset-A development sweep comprised 11 state counts across three load series, two forecasting horizons, two quantization locations, and three expanding-window folds, corresponding to 396 fold-specific evaluations. Based exclusively on development-stage evidence, the final state set was restricted to 3, 5, 7, and 16 states. The frozen final evaluation therefore comprised 48 Dataset-A configurations and 16 Dataset-B configurations, giving 64 structured configurations across the two co-primary datasets.

Primary conclusions were therefore based on the magnitude and direction of paired effects, agreement across both co-primary datasets and the four series, configuration-level support density, consistency across state-count, horizon, and quantization-location strata, and leave-one-series-out sensitivity analysis. Configuration-level counts were interpreted descriptively rather than as independent statistical replicates.

\section{Results}

\subsection{Aggressive low-state PTQ produces a reproducible accuracy penalty that QAT-FT largely avoids}

\begin{figure}[t]
    \centering
    \includegraphics[width=\textwidth]
    {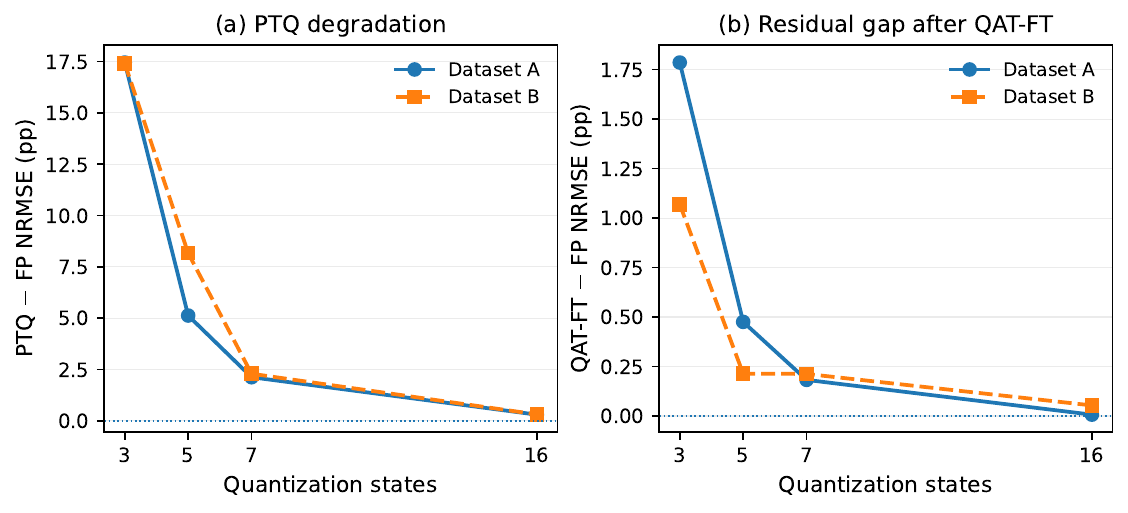}

    \caption{State-dependent PTQ degradation and QAT-FT residual gap relative to full precision. Mean NRMSE differences relative to the full-precision reference are shown for direct post-training quantization (PTQ) and quantization-aware fine-tuning (QAT-FT) at 3, 5, 7, and 16 weight states in Datasets A and B. Dataset-A state-level means summarize 12 configurations per state across three load series, two forecasting horizons, and two quantization locations; Dataset-B means summarize four configurations per state across one load series, two horizons, and two quantization locations. PTQ degradation is largest under the three-state constraint and decreases strongly as state resolution increases, whereas the residual QAT-FT difference from full precision remains substantially smaller across the tested state counts.}

    \label{fig:ptq_qat}
\end{figure}

Direct post-training quantization (PTQ) substantially degraded forecasting performance across the tested 3-, 5-, 7-, and 16-state weight representations. Across the 64-configuration structured comparison set spanning these state counts, both quantization locations, both forecasting horizons, and the load series within each co-primary dataset, mean NRMSE in Dataset A increased from 11.3299\% for the full-precision (FP) reference to 17.5770\% under PTQ, corresponding to a 6.2471-percentage-point penalty. In Dataset B, mean NRMSE similarly increased from 57.7067\% for FP to 64.7475\% under PTQ, a 7.0408-percentage-point penalty. QAT-FT was evaluated independently from the same pretrained FP starting point over the same 3-, 5-, 7-, and 16-state conditions and achieved substantially lower mean NRMSE values of 11.9424\% in Dataset A and 58.0938\% in Dataset B. Using the dataset-level mean NRMSE values, QAT-FT closed 90.20\% of the FP--PTQ performance gap in Dataset A and 94.50\% in Dataset B (Fig.~\ref{fig:ptq_qat}; Table~\ref{tab:headline_results}). \begin{table}[H]
\centering
\caption{Headline forecasting and accumulated-update results across the two co-primary datasets.}
\label{tab:headline_results}
\small
\begin{tabular}{lcc}
\hline
Metric & Dataset A & Dataset B \\
\hline
FP NRMSE (\%) & 11.3299 & 57.7067 \\
PTQ NRMSE (\%) & 17.5770 & 64.7475 \\
PTQ penalty vs.\ FP (pp) & +6.2471 & +7.0408 \\
QAT-FT NRMSE (\%) & 11.9424 & 58.0938 \\
PTQ penalty recovered by QAT-FT (\%) & 90.1957 & 94.5019 \\
Scratch $\alpha=0$ E40 NRMSE (\%) & 11.8001 & 57.6379 \\
Scratch high-$\alpha$ E40 NRMSE (\%) & 11.8417 & 57.8345 \\
High-$\alpha$ minus $\alpha=0$ NRMSE (pp) & +0.0416 & +0.1965 \\
Committed-update reduction (\%) & 99.7271 & 99.6741 \\
State-transition reduction (\%) & 79.9747 & 77.0203 \\
Commit acceptance (\%) & 0.2716 & 0.3247 \\
\hline
\end{tabular}
\end{table} Thus, across the full tested low-state range, direct quantization of the pretrained FP solution incurred a substantial average accuracy penalty, whereas independently applied QAT-FT produced models whose performance remained much closer to the FP reference.

The magnitude of the PTQ penalty depended strongly on the number of available weight states. At the most aggressive three-state condition, PTQ increased mean NRMSE relative to FP by 17.4513 percentage points in Dataset A and 17.4046 percentage points in Dataset B. The corresponding penalties decreased to 5.1237 and 8.1646 percentage points at five states, 2.1185 and 2.2876 percentage points at seven states, and only 0.2951 and 0.3063 percentage points at 16 states. QAT-FT remained substantially closer to FP at every tested state count. At three states, its residual mean NRMSE difference from FP was 1.7855 percentage points in Dataset A and 1.0687 percentage points in Dataset B; at 16 states, these differences decreased to only 0.0067 and 0.0534 percentage points, respectively. These results identify the three-state condition as the strongest quantization stress regime and show that the separation between PTQ and QAT-FT becomes particularly pronounced as the available weight-state representation becomes more restrictive (Fig.~\ref{fig:ptq_qat}).


The same overall pattern was broadly supported across the structured configuration comparisons. PTQ produced higher NRMSE than FP in 61 of 64 configurations (95.31\%), while QAT-FT produced lower NRMSE than PTQ in 62 of 64 configurations (96.88\%). Within Dataset A, QAT-FT outperformed the corresponding PTQ condition in all 48/48 matched cases, whereas in Dataset B it did so in 14/16 cases. The few non-supportive comparisons occurred in H1 FFN-Q conditions and are documented in the Supplementary Information. Collectively, the results show that direct PTQ becomes increasingly detrimental as the available weight-state representation becomes more restrictive, whereas independently evaluated QAT-FT provides substantially greater robustness to low-state quantization and closes most of the FP--PTQ performance gap in both co-primary datasets.
\subsection{Competitive low-state solutions can also be trained directly from scratch}

\begin{figure}[t]
    \centering
    \includegraphics[width=\textwidth]
    {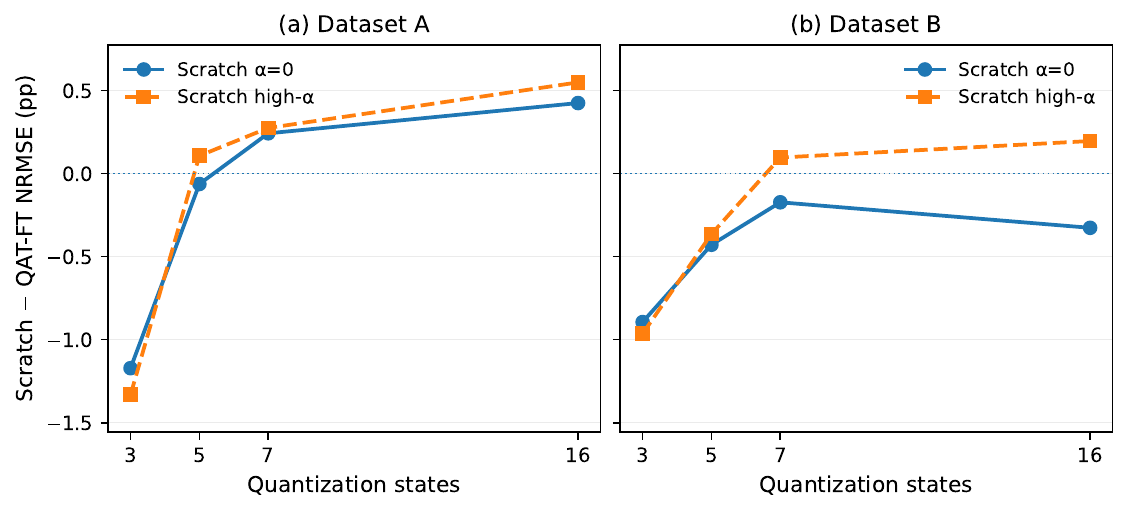}

    \caption{Direct Scratch-QAT relative to pretrained QAT-FT across low-state weight representations. State-dependent forecasting performance is compared for QAT-FT and the primary E40 Scratch-QAT conditions across 3, 5, 7, and 16 states in Datasets A and B. The clearest common advantage of direct Scratch-QAT occurs under the most restrictive three-state representation. At higher state counts, the relative behavior is less uniform, supporting the interpretation that Scratch-QAT is a competitive direct-training pathway rather than a generally superior replacement for pretrained QAT-FT. QAT-FT uses its frozen 25-epoch fine-tuning procedure, whereas E40 is the frozen primary Scratch-QAT endpoint.}

    \label{fig:scratch_qat}
\end{figure}

We next examined whether competitive low-state solutions could be obtained without initialization from a pretrained full-precision model. Both QAT-FT and Scratch-QAT use the same underlying quantization-aware formulation, in which continuous shadow weights are maintained while quantized weights are used in the forward pass; QAT-FT begins from a pretrained FP checkpoint, whereas Scratch-QAT begins from random Glorot-uniform initialization and follows its separately frozen training recipe. For the primary Scratch-QAT endpoint at 40 epochs (E40), averaged across the tested 3-, 5-, 7-, and 16-state conditions, both quantization locations, both forecasting horizons, and the load series within each dataset, Scratch \(\alpha=0\) achieved a mean NRMSE of 11.8001\% in Dataset A, compared with 11.9424\% for pretrained QAT-FT. In Dataset B, the corresponding values were 57.6379\% and 58.0938\%, respectively (Fig.~\ref{fig:scratch_qat}; Table~\ref{tab:headline_results}). Thus, the dataset-level E40 averages slightly favored Scratch-QAT in both datasets. However, this advantage was not uniform across individual configurations: Scratch \(\alpha=0\) E40 outperformed QAT-FT in 16/48 Dataset-A configurations and 13/16 Dataset-B configurations, corresponding to 29/64 (45.31\%) across the complete structured comparison set. These results therefore support the competitiveness of direct Scratch-QAT, rather than a general superiority over pretrained QAT-FT.

The strongest and most reproducible Scratch-QAT advantage occurred under the most restrictive three-state quantization. At the E40 endpoint, Scratch \(\alpha=0\) achieved a mean NRMSE 1.1717 percentage points lower than QAT-FT in Dataset A and 0.8942 percentage points lower in Dataset B. At the configuration level, Scratch-QAT was favorable in 9/12 three-state cases in Dataset A and 3/4 cases in Dataset B, giving 12/16 favorable three-state comparisons overall. Moreover, at the load-series summary level, three-state Scratch \(\alpha=0\) was favorable relative to QAT-FT across all four distinct load series nested within the two co-primary datasets. This concentration of support at three states is particularly notable because the same state count represented the strongest quantization-stress condition identified in Section 3.1 (Fig.~\ref{fig:ptq_qat}).

\begin{figure}[!t]
    \centering
    \includegraphics[width=\linewidth]
    {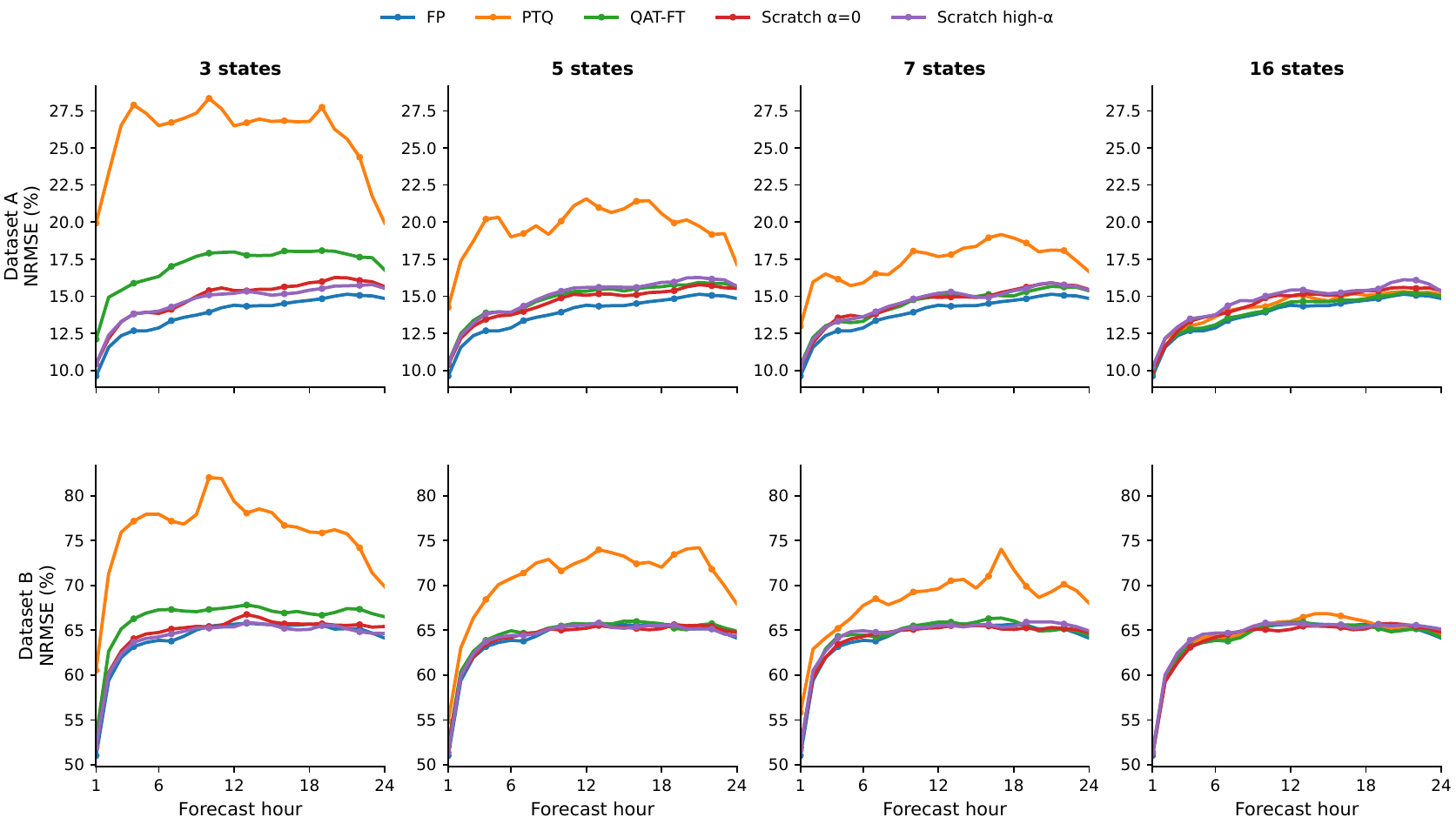}

    \caption{Forecast-lead dependence of H24 forecasting performance across quantization states and training strategies. NRMSE is shown separately for forecast hours 1--24. Columns correspond to 3, 5, 7, and 16 quantization states, with Dataset A shown in the upper row and Dataset B in the lower row. Curves show FP, PTQ, QAT-FT, primary E40 Scratch-QAT with $\alpha=0$, and E40 Scratch-QAT with the frozen high-$\alpha$ condition. For Dataset A, each quantized-method trajectory is the descriptive mean across the three load series and two quantization locations at the indicated state count; for Dataset B, each trajectory is the mean across the two quantization locations. FP is state independent and is shown as the corresponding dataset-level reference in each column.}
    
    \label{fig:h24_lead_dependence}
\end{figure}

To determine whether the favorable low-state Scratch-QAT behavior was confined to the aggregate H24 metric or persisted across the forecast horizon, NRMSE was examined separately for forecast hours 1--24 (Fig.~\ref{fig:h24_lead_dependence}). The three-state Scratch-QAT advantage extended across the full forecast horizon. In Dataset A, mean three-state Scratch $\alpha=0$ NRMSE changed from 10.4257\% at hour 1 to 15.6494\% at hour 24, compared with 12.0979\% to 16.7757\% for QAT-FT. In Dataset B, the corresponding Scratch trajectory changed from 51.9642\% to 65.4218\%, compared with 53.0955\% to 66.4997\% for QAT-FT. Thus, the mean three-state Scratch $\alpha=0$ trajectory remained lower than QAT-FT at all 24 forecast hours in both datasets.

High-$\alpha$ thresholding did not introduce a qualitatively different forecast-lead pattern. At three states, the high-$\alpha$ trajectory changed from 10.4327\% to 15.5368\% in Dataset A and from 51.9798\% to 64.6495\% in Dataset B, remaining close to the paired Scratch $\alpha=0$ trajectory throughout the horizon. Similar overlap was observed across the other state counts, consistent with the small aggregate accuracy changes accompanying the large activity reductions reported in Section~3.3. Overall, the per-hour analysis shows that the central low-state findings are not restricted to the aggregate H24 metric: PTQ degradation, QAT-FT recovery, the particularly strong three-state Scratch-QAT behavior, and the broadly preserved high-$\alpha$ forecasting trajectory are all visible across the 24 simultaneously predicted forecast hours.


Taken together, these results show that the quantization-aware training formulation can produce competitive low-state models from either a pretrained FP starting point or random initialization, although the relative advantage depends on the quantization regime. Scratch-QAT does not consistently outperform pretrained QAT-FT across the full configuration space; instead, its clearest reproducible advantage occurs under the most restrictive three-state condition. Accordingly, Scratch-QAT is best interpreted as a viable direct-training pathway for low-state models, rather than as a universally superior replacement for pretrained QAT-FT.
\subsection{Accumulated-update thresholding strongly suppresses committed and discrete target-state activity}

We next evaluated whether accumulated-update thresholding could reduce target-weight activity during Scratch-QAT while preserving forecasting performance. Scratch \(\alpha=0\) and Scratch high-\(\alpha\) were evaluated as exactly paired conditions, with only the accumulated-update threshold differing, as defined in Sections 2.6--2.7. Results are reported separately for committed shadow-weight updates and discrete quantized state transitions, which represent distinct levels of training activity.

\begin{figure}[H]
    \centering
    \includegraphics[width=\textwidth]
    {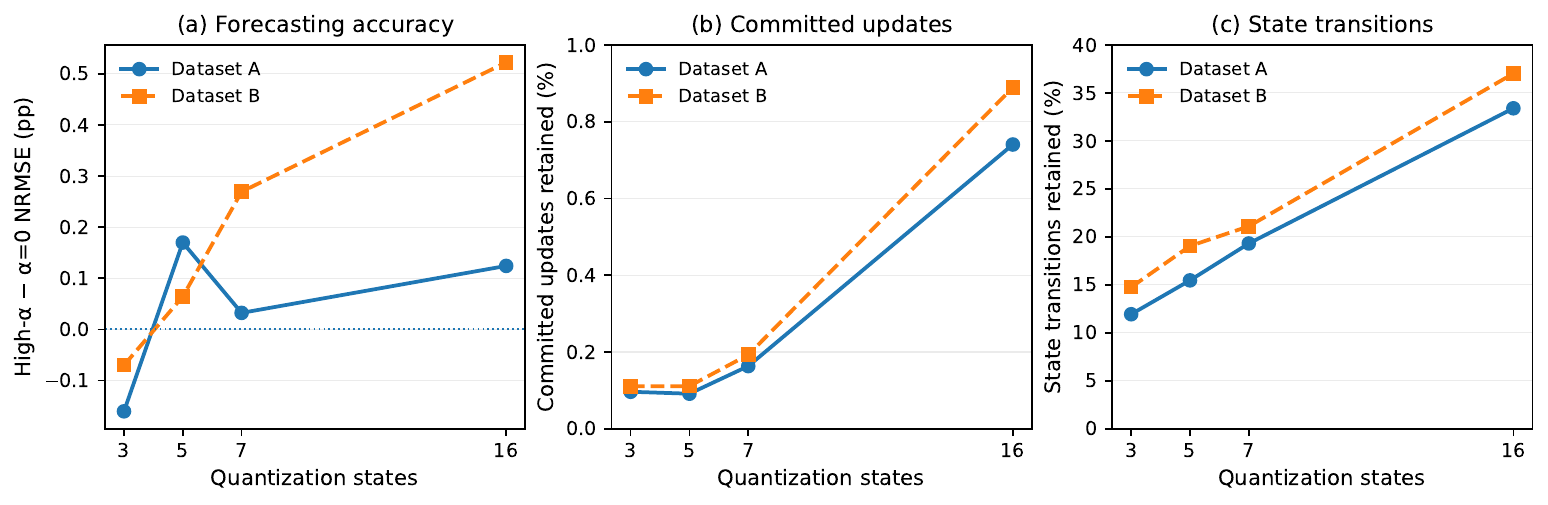}

    \caption{Forecasting accuracy and target-state activity under accumulated-update thresholding. Scratch high-\(\alpha\) is compared with its exactly paired Scratch \(\alpha=0\) control across the 3-, 5-, 7-, and 16-state conditions in both co-primary datasets. High-\(\alpha\) reduced committed shadow-weight updates by 99.7271\% in Dataset A and 99.6741\% in Dataset B, while discrete quantized state transitions were reduced by 79.9747\% and 77.0203\%, respectively. These large activity reductions were accompanied by mean NRMSE changes of only +0.0416 and +0.1965 percentage points. Committed-update suppression remained above 99\% at every tested state count in both datasets, whereas state-transition suppression remained substantial but decreased as the number of available states increased.}

    \label{fig:thresholding_activity}
\end{figure}

\begin{figure}[!t]
    \centering
    \includegraphics[width=\linewidth]
    {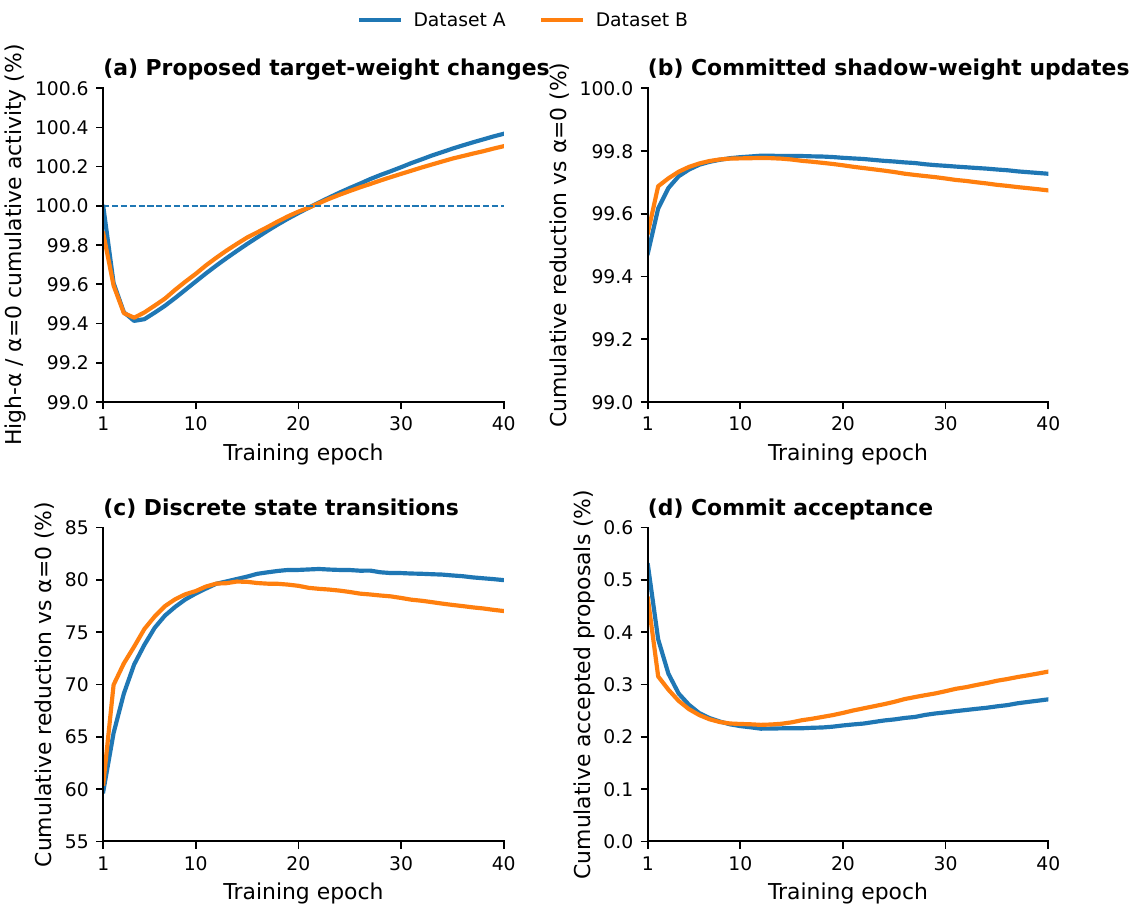}
    
    \caption{Epoch-wise evolution of accumulated-update thresholding activity during final Scratch-QAT training. Curves show the mean within-configuration cumulative quantities across the 48 Dataset-A and 16 Dataset-B paired $\alpha=0$/high-$\alpha$ final configurations over 40 epochs. (a) Cumulative optimizer-proposed target-weight changes under high-$\alpha$, expressed relative to the paired $\alpha=0$ control; values remain near 100\%, showing that the threshold does not suppress proposal generation. (b) Cumulative reduction in committed shadow-weight updates relative to $\alpha=0$. (c) Cumulative reduction in discrete quantized state-transition events relative to $\alpha=0$. (d) Cumulative commit-acceptance percentage under high-$\alpha$. Reductions were calculated within matched configurations before averaging so that architectures and quantization locations with different numbers of targeted parameters were not differentially weighted. These quantities describe algorithmic committed and discrete target-state activity; they do not represent measured physical memory writes, write energy, endurance, or hardware lifetime.}
    
    \label{fig:epochwise_activity}
\end{figure}

Across the tested 3-, 5-, 7-, and 16-state conditions, Scratch high-\(\alpha\) reduced committed shadow-weight updates by an average of 99.7271\% in Dataset A and 99.6741\% in Dataset B relative to the paired Scratch \(\alpha=0\) condition (Fig.~\ref{fig:thresholding_activity}; Table~\ref{tab:headline_results}). Mean commit acceptance rates were correspondingly only 0.2716\% and 0.3247\%, respectively, indicating that fewer than one-half of one percent of nonzero optimizer-proposed target-weight updates resulted in committed shadow-weight updates on average. These reductions in committed activity were accompanied by substantially fewer discrete quantized state transitions, with mean reductions of 79.9747\% in Dataset A and 77.0203\% in Dataset B. The transition reduction remained substantial at every tested state count: 88.07\%/85.24\% in Dataset A/B at three states, 84.54\%/80.95\% at five states, 80.69\%/78.90\% at seven states, and 66.61\%/62.98\% at 16 states. In contrast, committed shadow-weight update reduction remained above 99\% at every state count in both datasets, further demonstrating that committed shadow-weight updates and discrete state transitions represent different levels of training activity.

To examine whether this suppression persisted throughout training, the complete final Scratch-QAT histories were analyzed over the 40 training epochs (Fig.~\ref{fig:epochwise_activity}). Within each paired configuration, cumulative high-$\alpha$ activity was compared with its matched $\alpha=0$ control before averaging across configurations. Optimizer-proposed target-weight activity remained essentially unchanged, with the E40 ratio of cumulative proposed-update activity under high-$\alpha$ to that under $\alpha=0$ reaching 100.37\% in Dataset A and 100.30\% in Dataset B. In contrast, committed-update suppression reached 99.73\%/99.67\% at E40, while discrete state-transition suppression reached 79.97\%/77.02\%. Mean cumulative commit acceptance remained below 1\% throughout training and ended at 0.272\%/0.325\%. These trajectories confirm that thresholding persistently suppresses committed and discrete target-state activity rather than reducing optimizer-proposed activity.

These large activity reductions were accompanied by comparatively small changes in forecasting error. In Dataset A, mean NRMSE changed from 11.8001\% for Scratch \(\alpha=0\) to 11.8417\% for Scratch high-\(\alpha\), a difference of only +0.0416 percentage points. High-\(\alpha\) nevertheless achieved lower NRMSE than its paired \(\alpha=0\) run in 22/48 configurations, while 39/48 remained within +0.25 percentage points of the \(\alpha=0\) result. In Dataset B, mean NRMSE changed from 57.6379\% to 57.8345\%, corresponding to +0.1965 percentage points; high-\(\alpha\) achieved lower NRMSE in 6/16 configurations and remained within +0.25 percentage points in 9/16. Thus, orders-of-magnitude suppression of committed activity and large reductions in discrete state transitions were obtained without a comparably large change in mean forecasting performance.

The most aggressive three-state condition provides a particularly informative example of this trade-off. In Dataset A, Scratch high-\(\alpha\) reduced committed shadow-weight updates by 99.9036\% and discrete state transitions by 88.0657\% relative to Scratch \(\alpha=0\), while mean NRMSE was 0.1602 percentage points lower. Dataset B showed the same state-specific direction: committed shadow-weight updates fell by 99.8892\%, state transitions by 85.2441\%, and mean NRMSE was 0.0700 percentage points lower than under \(\alpha=0\).

Overall, accumulated-update thresholding reduced committed shadow-weight updates by approximately 99.7\% and discrete state transitions by approximately 77--80\% across the two datasets while maintaining comparable forecasting accuracy. These reductions concern committed and discrete target-state activity rather than fewer Adam steps or lower conventional training FLOPs, potentially making the approach relevant to low-state implementations in which physical state changes incur write costs; however, any hardware implications remain prospective.
\subsection{Representative forecast traces under frozen evaluation settings}

\begin{figure}[p]
    \centering
    \includegraphics[width=\linewidth]
    {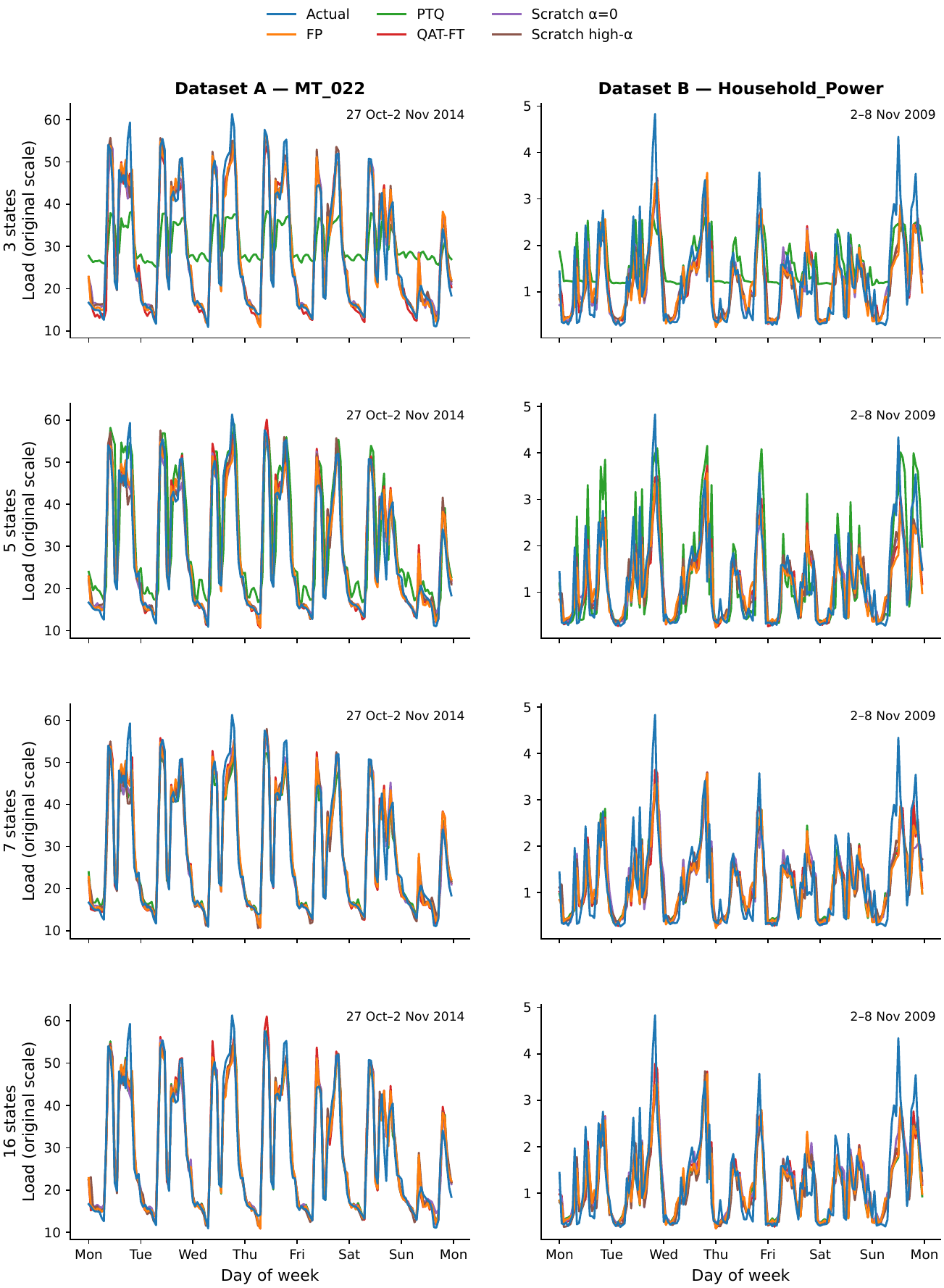}

    \caption{Representative H1 forecast traces across training strategies.}
    
    \label{fig:h1_representative_traces}
\end{figure}

\begin{figure}[p]
    \centering
    \includegraphics[width=\linewidth]
    {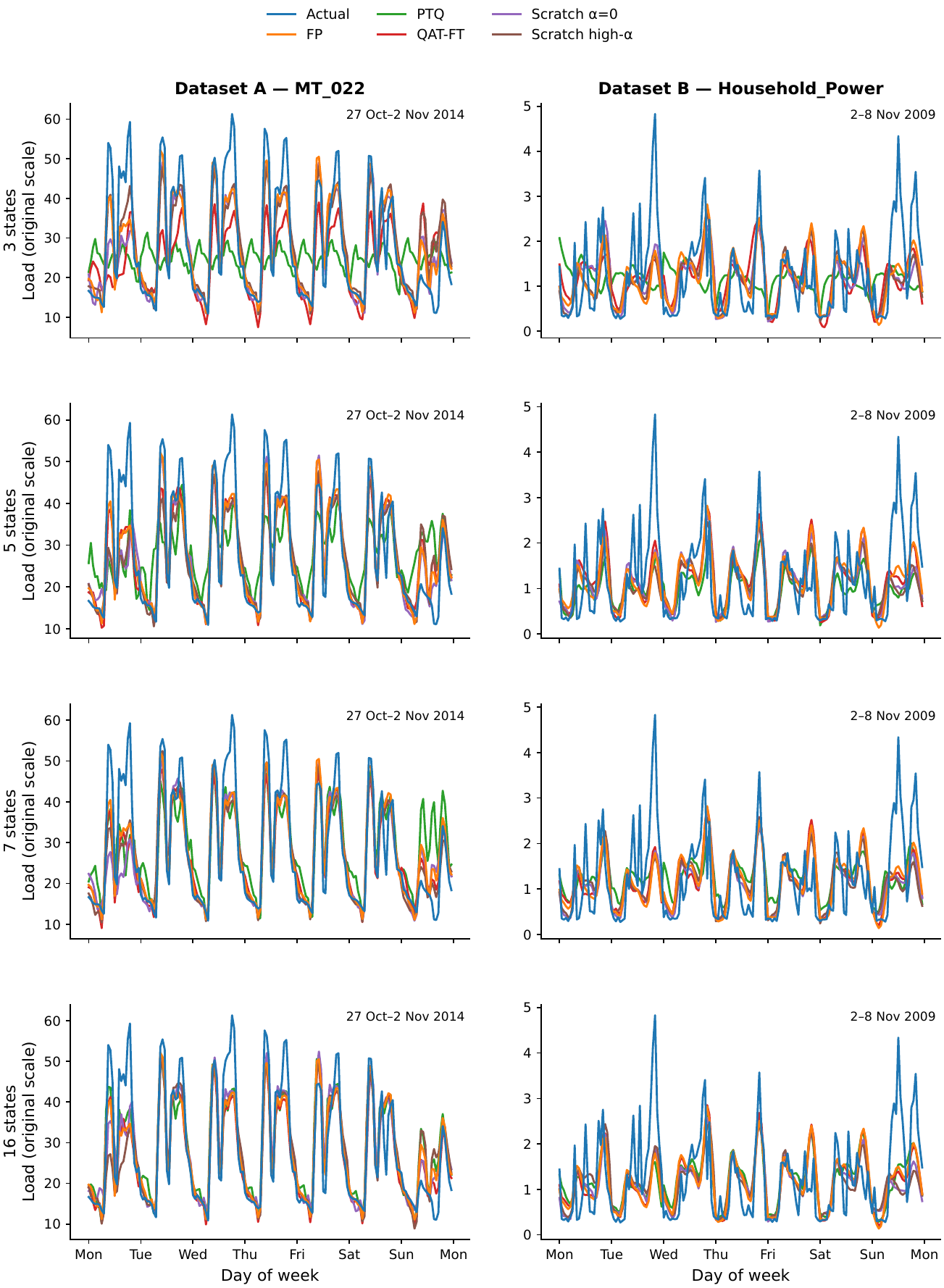}

    \caption{Representative H24 forecast traces across training strategies.}
    
    \label{fig:h24_representative_traces}   
\end{figure}

Representative forecast traces (Figs.~\ref{fig:h1_representative_traces} and~\ref{fig:h24_representative_traces}) are provided to visually complement the aggregate final-test analyses. Dataset A is represented by MT\_022 and Dataset B by Household\_Power. Quantized comparisons use the broader Attention+FFN-Q condition at 3, 5, 7, and 16 states and include PTQ, QAT-FT, E40 Scratch-QAT with $\alpha=0$, and E40 Scratch-QAT with the frozen high-$\alpha$ setting.

The displayed weeks were selected without reference to any quantized-model predictions. Complete Monday–Sunday weeks were first ranked by actual-load variability, and only the 50\% most variable weeks were retained; among these, the week whose FP H1 NRMSE was closest to the corresponding full final-test FP H1 NRMSE was selected. This yielded 27 October--2 November 2014 for MT\_022 and 2--8 November 2009 for Household\_Power. The same calendar weeks were used for the H24 visualization, where seven non-overlapping direct 24-h forecasts initiated at 00:00 each day were concatenated chronologically.

\subsection{Primary effects reproduce across datasets and load series}

The principal findings reproduced across the two co-primary datasets under the methodology transferred from Dataset A without Dataset-B-specific retuning. At the series-summary level, all four distinct load series retained the expected directions for PTQ degradation, QAT-FT improvement over PTQ, committed-update suppression, and discrete state-transition suppression (Table~\ref{tab:series_robustness}).

\begin{table}[H]
\centering
\caption{Series-level mean effects across the four load series.}
\label{tab:series_robustness}
\small
\setlength{\tabcolsep}{4pt}
\begin{tabular}{llrrrr}
\toprule
Dataset & Series &
\shortstack{PTQ $-$ FP\\(pp)} &
\shortstack{QAT-FT $-$ PTQ\\(pp)} &
\shortstack{Commit\\reduction (\%)} &
\shortstack{Transition\\reduction (\%)} \\
\midrule
Dataset A & MT\_004          & +7.4708 & -7.0838 & 99.7391 & 80.5192 \\
Dataset A & MT\_022          & +7.2828 & -5.9642 & 99.6379 & 76.6080 \\
Dataset A & MT\_324          & +3.9878 & -3.8559 & 99.8044 & 82.7968 \\
Dataset B & Household\_Power & +7.0408 & -6.6537 & 99.6741 & 77.0203 \\
\bottomrule
\end{tabular}
\end{table}

Directional consistency was also retained after stratification by each principal experimental factor. In both datasets, all four primary effects retained their expected direction at every tested state count, both forecasting horizons, and both quantization locations (Table~\ref{tab:factor_concordance}).

\begin{table}[H]
\centering
\caption{Cross-dataset directional concordance across the principal experimental factors.}
\label{tab:factor_concordance}
\small
\setlength{\tabcolsep}{4pt}
\begin{tabular}{lccccc}
\toprule
Factor &
Levels &
\shortstack{PTQ\\degradation} &
\shortstack{QAT-FT improvement\\over PTQ} &
\shortstack{Commit\\reduction} &
\shortstack{Transition\\reduction} \\
\midrule
State count           & 4 & 4/4 & 4/4 & 4/4 & 4/4 \\
Forecast horizon      & 2 & 2/2 & 2/2 & 2/2 & 2/2 \\
Quantization location & 2 & 2/2 & 2/2 & 2/2 & 2/2 \\
\bottomrule
\end{tabular}
\end{table}

Leave-one-series-out sensitivity analysis further tested whether any principal conclusion depended disproportionately on a single load series (Table~\ref{tab:loo_robustness}).

\begin{table}[H]
\centering
\caption{Leave-one-series-out sensitivity of the four primary effects.}
\label{tab:loo_robustness}
\small
\setlength{\tabcolsep}{4pt}
\begin{tabular}{lcccc}
\toprule
Primary effect &
\shortstack{Expected\\direction} &
\shortstack{Four-series\\mean} &
\shortstack{LOO mean\\range} &
\shortstack{All four LOO analyses\\retain direction} \\
\midrule
PTQ degradation vs FP (pp)
& Positive & +6.4455 & +6.1038 to +7.2648 & Yes \\

QAT-FT improvement vs PTQ (pp)
& Negative & -5.8894 & -6.5672 to -5.4913 & Yes \\

High-$\alpha$ commit reduction (\%)
& Positive & 99.7139 & 99.6837 to 99.7392 & Yes \\

High-$\alpha$ transition reduction (\%)
& Positive & 79.2361 & 78.0492 to 80.1121 & Yes \\
\bottomrule
\end{tabular}
\end{table}

Every primary conclusion retained its expected direction after removal of MT\_004, MT\_022, MT\_324, or Household\_Power individually. The distinctive three-state Scratch-QAT result was also consistent across the four series: at the series-summary level, Scratch-QAT $\alpha=0$ achieved lower mean NRMSE than QAT-FT for all four distinct load series. Together, these results show that the principal low-state effects were preserved across datasets, load series, state counts, forecasting horizons, and quantization locations and were not dependent on any single series.

\section{Discussion}
\subsection{Why aggressive PTQ fails most severely at very low state counts}

The strong state dependence of the PTQ penalty is consistent with the increasingly coarse representation imposed as the number of available weight states decreases. Because the FP model is optimized in a continuous parameter space and only quantized after training, aggressive PTQ can displace learned weights from their optimized values without allowing the model to adapt to the discrete constraint. As the state count increases, the quantization grid becomes finer and this mismatch is reduced, consistent with the progressive approach toward FP performance observed at higher state counts. This interpretation agrees with prior studies showing that direct low-precision post-training quantization can degrade model accuracy, particularly under restrictive numerical representations \citep{Krishnamoorthi2018,Nagel2019}.

\subsection{Why QAT-FT is more robust than direct PTQ}

The greater robustness of QAT-FT is consistent with its ability to optimize the model while the targeted weights are already subject to the low-state representation during the forward pass. Rather than projecting an FP solution onto a discrete grid only after optimization, QAT-FT allows the continuous shadow weights to adapt under the errors introduced by quantization, enabling the learned solution to accommodate the available weight states. This distinction becomes particularly important under aggressive quantization, where the mismatch between an unconstrained FP solution and its discrete representation is largest. Similar benefits of incorporating quantization into the training process have been reported in prior low-precision neural-network studies \citep{Hubara2018,Jacob2018}.

\subsection{What Scratch-QAT adds}

Scratch-QAT addresses a different question from QAT-FT: rather than adapting a pretrained FP solution to a low-state representation, it tests whether a useful solution can emerge directly while quantization is active throughout training. The fact that competitive models could be obtained from random initialization therefore indicates that full-precision pretraining is not strictly required for learning under severe weight-state constraints. The particularly favorable behavior observed at three states is notable because this is also the regime in which direct PTQ was most disruptive, suggesting that learning directly within the constrained representation can be especially valuable when the available state space is extremely limited. On average, Scratch \(\alpha=0\) achieved slightly lower NRMSE than QAT-FT in both datasets, by 0.1423 and 0.4559 percentage points in Datasets A and B, respectively. However, because the advantage was not consistent across all state counts and configurations, Scratch-QAT should be interpreted as a viable alternative training pathway without requiring full-precision pretraining, rather than a generally superior replacement for pretrained QAT-FT.

\subsection{Accumulated-update thresholding changes the relevant notion of training activity}

A key implication of accumulated-update thresholding is that optimizer activity and physically meaningful state-changing activity need not be equivalent. Conventional training treats each optimizer update as part of the computational workload, whereas a low-state implementation may be more sensitive to how often those updates ultimately require a committed weight change or a transition between discrete states. The strong suppression of such activity with little change in forecasting performance suggests that many small optimizer-driven adjustments are not individually necessary for maintaining an effective discrete solution. Accumulating these adjustments before committing them therefore provides a form of temporal filtering of parameter updates, potentially allowing the optimization process to retain gradual information while avoiding unnecessary state changes. This behavior could be particularly advantageous for hardware implementations based on multi-state synaptic devices, where changing a stored weight may require a physical programming operation. The observed reduction in committed and discrete target-state activity is therefore potentially relevant to implementations in which state changes incur physical programming costs; quantifying any resulting effects on write activity, latency, or endurance will require device-specific validation beyond the present study.

\section{Conclusion}
This study shows that extremely low-state Transformer learning remains viable for electricity-load forecasting, but that the training pathway becomes critical as the available weight representation becomes increasingly discrete. Direct PTQ produced substantial degradation at the lowest state counts, whereas QAT-FT recovered most of the resulting accuracy loss across both co-primary datasets. Scratch-QAT further demonstrated that competitive low-state solutions can emerge directly from random initialization without requiring a pretrained full-precision precursor, with particularly favorable behavior in the three-state regime. Accumulated-update thresholding provided a complementary reduction in training activity by allowing small optimizer-proposed changes to accumulate before being committed, reducing committed shadow-weight updates by approximately 99.7\% and discrete state transitions by approximately 77--80\% while producing only small mean changes in forecasting error. The principal PTQ, QAT-FT, and thresholding effects remained consistent under transfer of the frozen methodology to the second co-primary dataset without dataset-specific retuning. Together, these findings show that aggressive weight-state reduction, direct low-state learning, and strong suppression of committed state-changing activity can coexist within a Transformer forecasting framework. Looking forward, AI hardware in which nanoscale multilevel nonvolatile devices directly encode highly quantized weights and are integrated with CMOS neuron and peripheral circuitry could provide a natural architecture for translating such low-state learning into compact, high-performance, and energy-efficient computing systems \citep{legallo2018mixedprecision,sebastian2020memory,nandakumar2020mixedprecision,Misba2022,alam2024quantizedautoencoder,Aguirre2024,dhull2025quantizeddw}.

\section{Acknowledgments}
This research was supported by the Commonwealth Cyber Initiative (CCI) under Grant No. VV-1Q23-008.


\bibliographystyle{plainnat}
\bibliography{references}

\end{document}